\documentclass[11pt]{article}

\usepackage[final]{acl}

\usepackage{times}
\usepackage{latexsym}

\usepackage[T1]{fontenc}

\usepackage[utf8]{inputenc}

\usepackage{microtype}

\usepackage{inconsolata}

\usepackage{graphicx}

\usepackage{amsfonts}
\usepackage{tikz}
\usetikzlibrary{positioning, arrows.meta}

\usepackage{subcaption}
\usepackage{multirow}
\usepackage{algorithm}
\usepackage{algpseudocode}
\usepackage{amsmath}
\usepackage{amsthm}
\usepackage{bbm}
\usepackage{amssymb}

\usepackage{booktabs}

\usepackage{graphicx}
\usepackage{subcaption} 

\definecolor{myblue}{HTML}{0022ff}  

\theoremstyle{plain}

\theoremstyle{definition}

\theoremstyle{remark}

\DeclareMathOperator{\softmax}{softmax}

\usepackage[nameinlink,capitalise]{cleveref}

\newcommand\penn{$^\dagger$}
\newcommand\uiuc{$^\ddagger$}
\newcommand\oist{$^\S$}

\title{When LRP Diverges from Leave-One-Out in Transformers}

\author{ \\
\textbf{Weiqiu You}\penn{} \quad \textbf{Siqi Zeng}\uiuc{} \quad \textbf{Yao-Hung Hubert Tsai}\oist{} \quad \textbf{Makoto Yamada}\oist{}  \quad \textbf{Han Zhao}\uiuc{} \\[2ex]
   \penn{}University of Pennsylvania, Philadelphia, PA, USA \\
   \uiuc{}University of Illinois Urbana-Champaign, Urbana, IL, USA \\
   \oist{}Okinawa Institute of Science and Technology, Okinawa, Japan
   }

\begin{document}
\maketitle

\begin{abstract}
Leave-One-Out (LOO) provides an intuitive measure of feature importance but is computationally prohibitive.
While Layer-Wise Relevance Propagation (LRP) offers a potentially efficient alternative, its axiomatic soundness in modern Transformers remains largely under-examined.
In this work, we first show that the bilinear propagation rules used in recent advances of AttnLRP violate the implementation invariance axiom. 
We prove this analytically and confirm it empirically in linear attention layers.
Second, we also revisit CP-LRP as a diagnostic baseline and find that bypassing relevance propagation through the softmax layer---backpropagating relevance only through the value matrices---significantly improves alignment with LOO, particularly in middle-to-late Transformer layers.
Overall, our results suggest that (i) bilinear factorization sensitivity and (ii) softmax propagation error potentially jointly undermine LRP’s ability to approximate LOO in Transformers.~\footnote{Correspondence to \texttt{weiqiuy@seas.upenn.edu}. Code is available at \url{https://github.com/fallcat/attn_loo}.
}

\end{abstract}

\section{Introduction}
\label{sec:intro}

\begin{figure}[t]
\centering
\begin{tikzpicture}[
  node distance=1.4cm and 1.5cm,
  every node/.style={font=\small},
  input/.style={draw, circle, draw=black!70, fill={rgb,255:red,224; green,235; blue,255}},
  hidden/.style={draw, circle, draw=black!70, fill={rgb,255:red,240; green,240; blue,240}},
  output/.style={draw, circle, draw=black!70, fill={rgb,255:red,220; green,245; blue,230}},
  arrow/.style={-{Latex}, thick},
  dashedarrow/.style={-{Latex}, dashed, thick}
]

\node[input] (x1a) {\(x_1\)};
\node[input, below=1cm of x1a] (x2a) {\(x_2\)};
\node[input, below=1cm of x2a] (x3a) {\(x_3\)};
\node[hidden, right=1cm of x2a] (h1a) {\(h = x_1 x_2\)};
\node[output, above=0.7cm of h1a, right=1cm of h1a] (ya) {\(y = hx_3\)};

\draw[arrow] (x1a) -- (h1a);
\draw[arrow] (x2a) -- (h1a);
\draw[arrow] (h1a) -- (ya);
\draw[arrow] (x3a) -- (ya);

\node[align=center, above=1.2cm of h1a] {\textbf{Network A}\\Left-associative};
\node[align=center, below=0.2cm of x1a, xshift=-0.5cm] {\(x_1=2, R_{x_1}=6\)};
\node[align=center, below=0.2cm of x2a, xshift=-0.5cm] {\(x_2=3, R_{x_2}=6\)};
\node[align=center, below=0.2cm of x3a, xshift=-0.5cm] {\(x_3=4, R_{x_3}=12\)};

\node[input, below=5cm of x1a] (x1b) {\(x_1\)};
\node[input, below=1cm of x1b] (x2b) {\(x_2\)};
\node[input, below=1cm of x2b] (x3b) {\(x_3\)};
\node[hidden, right=1cm of x2b] (h1b) {\(h' = x_2 x_3\)};
\node[output, above=0.7cm of h1b, right=1cm of h1b] (yb) {\(y = x_1 h'\)};

\draw[arrow] (x2b) -- (h1b);
\draw[arrow] (x3b) -- (h1b);
\draw[arrow] (h1b) -- (yb);
\draw[arrow] (x1b) -- (yb);

\node[align=center, above=1.2cm of h1b] {\textbf{Network B}\\Right-associative};
\node[align=center, below=0.2cm of x1b, xshift=-0.5cm] {\(x_1=2, R_{x_1}=12\)};
\node[align=center, below=0.2cm of x2b, xshift=-0.5cm] {\(x_2=3, R_{x_2}=6\)};
\node[align=center, below=0.2cm of x3b, xshift=-0.5cm] {\(x_3=4, R_{x_3}=6\)};

\end{tikzpicture}
\caption{Two functionally equivalent network factorizations of the same function \(y = x_1 x_2 x_3\). Despite identical outputs, LRP's $\varepsilon$-rule assigns different relevance to inputs depending on factorization.}
\label{fig:bilinear_diagram}
\end{figure}
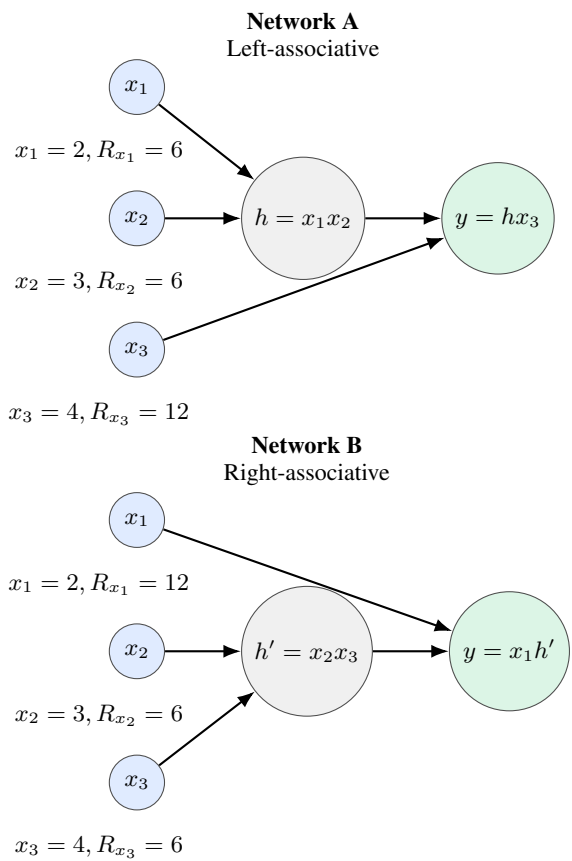

As Transformer-based machine learning models become central to high-stakes domains like healthcare~\citep{DBLP:journals/corr/abs-1907-07374,Hameed2023what} and legal systems~\citep{Zeng_2016,wexler2017computer}, the need for faithful explanations is critical. 
In particular, unfaithful explanations risk misleading domain experts, thereby undermining their ability to make informed decisions.

An intuitive but computationally prohibitive explanation method is Leave-One-Out (LOO), which measures the impact of removing each feature on a model's prediction. To approximate LOO efficiently, AttentionRollout~\citep{abnar-zuidema-2020-quantifying} was introduced to move beyond simplistic single-layer attention scores by composing them across layers, offering a more intuitive way to track information flow. However, these methods are incomplete as they propagate only attention scores while ignoring other key components like values and hidden states. 

Layer-Wise Relevance Propagation (LRP)~\citep{Bach2015LRP} redistributes a model's output score backward through the network while conserving relevance across layers, offering a more principled framework for approximating LOO.
Recent advances in LRP for attention rules in Transformers~\citep{vaswani2017attention} include CP-LRP~\citep{Ali22cplrp} and AttnLRP~\citep{Achtibat2024AttnLRP}.
CP-LRP regards the matrix multiplication between attention weights and value as a linear layer, propagating relevance scores only through the value vectors~\citep{Ali22cplrp}.
AttnLRP critiques CP-LRP for being unfaithful by ignoring the propagation through the attention weights, and proposes a new rule for bilinear layer using deep taylor decomposition (DTD), which improves upon insertion and deletion related metrics~\citep{Achtibat2024AttnLRP}.

However, previous works have shown that LRP in general violates \textit{implementation invariance} axiom---the principle that functionally identical networks should yield identical explanations~\citep{Sundararajan17Axiomatic}.
This flaw was first shown in a counterexample where two networks, differing only in their internal arrangement of ReLU operations, produced different LRP attributions despite being functionally equivalent~\citep{Sundararajan17Axiomatic}.
With the recent CP-LRP and AttnLRP, it has not yet been shown whether this flaw still exists or whether these LRP variants can approximate LOO in Transformers well despite these limitations.

In this work, we provide the first formal proof that LRP's axiomatic failure is not a rare corner case but still exists in AttnLRP's new rule for bilinear layers.
We establish this through both theory and experiments: first, by proving the violation with a simple analytical example, and then by empirically demonstrating it in a one-layer linear attention model.
Moreover, we empirically find that, although CP-LRP lags behind AttnLRP in insertion and deletion based metrics, CP-LRP is actually a better approximation for LOO.
Ablating their application layer by layer, we find that treating attention weights as constants and bypassing softmax-layer backpropagation (as in CP-LRP) improves LOO alignment in middle-to-late layers, whereas AttnLRP’s bilinear propagation can be more beneficial in earlier layers.
This is observed in our layerwise analysis (\Cref{sec:experiments}), where several middle-to-late layers exhibit flatter attention distributions.

Our contributions are twofold and relate to different parts of attention mechanisms:

\paragraph{Error in Bilinear Layers.} To the best of our knowledge, we are the first to formally prove that AttnLRP’s propagation rule for bilinear layers violates the implementation invariance axiom. We identify a fundamental source of LOO estimation error that extends beyond previously studied nonlinearities to the core bilinear operations underlying modern attention mechanisms. Using a small linear attention model trained on MNIST~\citep{lecun1998mnist}, we empirically confirm that LRP attributions differ between left- and right-associative bilinear factorizations: their scores are not fully correlated and exhibit inconsistent alignment with LOO.

\paragraph{Error in Softmax Layers.} 
We compare the two Transformer-based LRP variants, CP-LRP (LRP for Transformers without softmax propagation) and AttnLRP (LRP for Transformers with bilinear and softmax propagation), and show that CP-LRP correlates with LOO better in BERT~\citep{Devlin2019BERT} on SST~\citep{Socher2013Recursive} and IMDB~\citep{Maas2011Learning}.
We conduct a layer-wise ablation and show that regarding the attention weights as constant like CP-LRP in the middle-to-late layers in Transformers is the most helpful for approximating LOO.

\section{Background}
\label{sec:background}

In this section, we build the foundation for our analysis by first establishing the desired properties of an attribution method and then detailing the mechanics of LRP as a popular approximation. We begin by formally defining LOO attribution, establishing it as a conceptual benchmark due to its faithfulness and implementation invariance.
As LOO has a prohibitive computational cost, there is a pressing need for efficient alternatives. We then introduce LRP, a method designed to overcome the limitations of both LOO and early heuristics that only requires one forward pass of models. We will describe its core principles of relevance conservation and layer-wise decomposition before examining its specific adaptations for the Transformer architecture, setting the stage for our theoretical critique.

\subsection{Leave-One-Out (LOO) Attribution}

A central challenge in attribution evaluation is identifying a reliable reference metric that reflects the true contribution of each input feature. The LOO score provides such a reference by measuring the change in the model's output when a single feature is removed. Formally, for an input $x$ and feature $i$, the LOO score is calculated as:
\begin{equation}
\text{LOO}_i = f(x) - f(x_{\setminus i})
\label{eqn:loo}
\tag{LOO}
\end{equation}
where $x_{\setminus i}$ denotes the input with feature $i$ removed. Because this definition depends only on the model’s \textbf{functional behavior}, the abstract rule that defines an operation as a mapping from inputs to outputs---rather than its specific implementation---the particular procedure to realize that mapping, LOO, is invariant to specific implementations and satisfies the \textbf{implementation invariance} axiom. 
LOO is widely regarded as a common baseline for feature importance~\citep{Ancona2019Towards, Covert2021Explaining}.
However, computing LOO scores exactly requires one forward pass per feature, which is computationally prohibitive for modern deep models. This motivates the development of scalable attribution methods that aim to approximate LOO scores efficiently while preserving their desirable properties.

\subsection{Heuristic LOO approximations}
Early attempts to approximate LOO in Transformers focused on the attention mechanism itself, as attention weights provide an intuitive story for how information is combined across the input. Methods like Attention Rollout \citep{abnar-zuidema-2020-quantifying} sought to address the limitation of using raw, single-layer attention weights by composing them across layers. The intuition is to recursively propagate attention scores from the final layer to the input. The effective attention from layer $l$ down to the input is computed by recursively multiplying attention matrices:
\begin{equation}
\tilde{A}^{(l)} = A^{(l)} \cdot \tilde{A}^{(l-1)}
\label{eqn:rollout}
\tag{Rollout}
\end{equation}
where $A^{(l)}$ is the attention matrix at layer $l$ and the rollout begins with an identity matrix.

However, such heuristic methods are incomplete approximations of LOO. By focusing exclusively on attention matrices ($A$), they ignore the contributions of other critical components, such as the value projections ($V$), feed-forward networks, and residual connections. The true final prediction is a function of the entire computation path, not just the attention patterns. The significance of this limitation becomes clear when considering findings that Transformer performance can be surprisingly robust even when learned attention is replaced with hard-coded, non-data-dependent patterns~\citep{you-etal-2020-hard}. This suggests that much of the model's predictive power is encoded in the value transformations and subsequent layers---a significant part of the model that attention-only methods completely disregard.

\subsection{A Principled LOO Alternative: Layer-Wise Relevance Propagation (LRP)}
\label{sec:loo_lrp}
To overcome the shortcomings of heuristic methods, more principled approaches like LRP have been adapted for Transformers. 
LRP offers a complete decomposition of the model's prediction by propagating relevance backward from the output to the input layer-by-layer.
It operates on a conservation principle, where the model's output score $f(x)$ is decomposed into a sum of relevance scores $R_i$ for each input feature, such that $f(x) = \sum_i R_i$. 
This is achieved by propagating the total relevance backward through the network, conserving it at each layer. 
In purely linear networks, this relevance redistribution is mathematically equivalent to computing Leave-One-Out (LOO) scores, 
making LRP a natural, computationally efficient approximation to LOO in more complex architectures that include nonlinear or multiplicative operations.

\paragraph{Epsilon Rule for Linear Layers.} 
For standard linear layers, a common propagation choice is the $\varepsilon$-rule, 
which distributes relevance from a neuron to its inputs in proportion to their contribution to its activation:
\begin{equation}
R^{(l-1)}_i 
= \sum_j 
\frac{z^{\mathrm{lin}}_{ij}}
{\sum_k z^{\mathrm{lin}}_{kj} + \varepsilon \cdot \mathrm{sign}\left(\sum_k z^{\mathrm{lin}}_{kj}\right)} 
\, R^{(l)}_j,
\label{eq:lrp-epsilon}
\end{equation}
where $z^{\mathrm{lin}}_{ij} = x^{(l-1)}_i W_{ij}$ represents the contribution of input neuron $i$ to neuron $j$. 
A small stabilizer $\varepsilon > 0$ is added in the denominator to prevent division by zero 
when the sum of input contributions $\sum_k z^{\mathrm{lin}}_{kj}$ is close to zero and to ensure numerical stability.
In this linear case where $f(x) = W x$, the resulting relevance assignments $R_i = x_i W_i$ 
exactly match the LOO scores.

However, commonly used networks nowadays contain nonlinear and multiplicative components 
(e.g., ReLU, bilinear attention, and softmax), for which this linear assumption no longer holds. 
To extend LRP beyond the linear case, the Deep Taylor Decomposition (DTD) framework~\citep{Montavon2017DTD} 
locally approximates nonlinear functions by their first-order Taylor expansion around a reference point. 
Relevance is then redistributed according to each input's contribution in this linearized neighborhood, 
ensuring local conservation but introducing potential approximation errors when activations interact nonlinearly. 
This DTD principle underlies the propagation rules that follow for bilinear and softmax layers.

\paragraph{Rules for Bilinear Layers.}
We make the attention forward pass explicit:
\begin{equation}
\begin{aligned}
Z&=\frac{1}{\sqrt{d_k}}\,QK^\top,\\
A&=\softmax(Z)\ \text{(row-wise)},\\
O&=AV.
\label{eq:attn-forward}
\end{aligned}
\end{equation}
Bilinearity exists in two places: \(O=AV\) and \(Z=QK^\top/\sqrt{d_k}\).

\textbf{CP-LRP}~\citep{Ali22cplrp} treats \(O=AV\) as linear in \(V\) (holding \(A\) fixed), backpropagating relevance only through \(V\) while \(A\) (and $Q,K$ as well) gets zero relevance scores.
\textbf{AttnLRP}~\citep{Achtibat2024AttnLRP} instead uses DTD~\citep{Montavon2017DTD} to split relevance between both factors. With a small stabilizer \(\varepsilon>0\):

Incoming relevance at \(O_{jp}\) (denoted \(R^{(l)}_{jp}\)) is split to \(A\) and \(V\) as
\begin{subequations}\label{eq:attn-bilinear}
\begin{align}
R^{(l-1)}_{A_{ji}}
&= \sum_{p}
\frac{A_{ji}\,V_{ip}}{\,2\,O_{jp}+\varepsilon\,\mathrm{sign}(O_{jp})\,}\; R^{(l)}_{jp},
\label{eq:attn-bilinear-A}\\
R^{(l-1)}_{V_{ip}}
&= \sum_{j}
\frac{A_{ji}\,V_{ip}}{\,2\,O_{jp}+\varepsilon\,\mathrm{sign}(O_{jp})\,}\; R^{(l)}_{jp}.
\label{eq:attn-bilinear-V}
\end{align}
\end{subequations}
Relevance is split in half, assigning the same value to two values that multiply together inside matrices in bilinear operations. 
The sum of relevance in attention $A$ is the same as the sum in value $V$: $R^{(l-1)}_A = R_V^{(l-1)} = \frac{1}{2}R_O^{(l)}$, in the limit $\varepsilon\rightarrow 0$.

Similarly, the relevance score from the pre-softmax attention logits $Z=QK^\top/\sqrt{d_k}$, where we define \(c_{jir}\!\triangleq\!\frac{Q_{jr}K_{ir}}{\sqrt{d_k}}\) so that \(Z_{ji}=\sum_r c_{jir}\), can be split to $Q$ and $K$ as
\begin{subequations}\label{eq:score-bilinear}
\begin{align}
R^{(l-1)}_{Q_{jr}}
&= \sum_{i}
\frac{c_{jir}}{\,2\,Z_{ji}+\varepsilon\,\mathrm{sign}(Z_{ji})\,}\; R^{(l)}_{Z_{ji}},
\label{eq:score-bilinear-Q}\\
R^{(l-1)}_{K_{ir}}
&= \sum_{j}
\frac{c_{jir}}{\,2\,Z_{ji}+\varepsilon\,\mathrm{sign}(Z_{ji})\,}\; R^{(l)}_{Z_{ji}}.
\label{eq:score-bilinear-K}
\end{align}
\end{subequations}
The sum of relevance in query $Q$ is the same as the sum in key $K$: $R^{(l-1)}_Q = R_K^{(l-1)} = \frac{1}{2}R_Z^{(l)}$, in the limit $\varepsilon\rightarrow 0$.


These AttnLRP bilinear rules are used with the softmax rule in the next subsection. For the whole backward propagation process, 
relevance is first split at \(O\!\to\!(A,V)\) via \Cref{eq:attn-bilinear},
then passed through \(A=\softmax(Z)\) (\Cref{eq:softmax-lrp}),
and finally split at \(Z\!\to\!(Q,K)\) via \Cref{eq:score-bilinear}.

\paragraph{Rule for the Softmax Layer.}
AttnLRP also derives rules for softmax layers using DTD. Let $Z$ be the logits and $A=\softmax(Z)$ the attention weights. 
We write superscripts $(l),(l\!-\!1)$ for the layers indicating propagation direction and use subscripts to denote tensors ($Z,A,V$).
Following \citet{Achtibat2024AttnLRP}, a first-order Taylor expansion leads to the element-wise rule
\begin{equation}
\begin{aligned}
R^{(l-1)}_{Z_{ji}}
\;=\;
Z_{ji}\!\left( 
R^{(l)}_{A_{ji}}
-
A_{ji} \sum\nolimits_{i'} R^{(l)}_{A_{ji'}}
\right),
\\
A_{ji} = \softmax(Z_{j:})_i,
\label{eq:softmax-lrp}
\end{aligned}
\end{equation}
which is applied entry-wise over $(j,i)$ to $Z$ and $A$.
Rule \Cref{eq:softmax-lrp} moves relevance from $A$ back to the logits $Z$; from there, the bilinear rule for $Z$ distributes relevance to $Q$ and $K$ (see \Cref{eq:score-bilinear}).
Together with \Cref{eq:attn-bilinear}, these rules conserve relevance locally while making explicit which components (attention weights vs.\ values vs.\ logits) carry the propagated mass.

Since CP-LRP treats bilinear layers as linear, relevance is not propagated through \(A\), and consequently, the softmax propagation rule— which depends on relevance scores from the attention—does not apply.

While these rules are derived from a principled framework and satisfy local relevance conservation, they rely on intermediate activations from the forward pass (e.g. $z_{ij}$ in the $\varepsilon$-rule, which is computed using the input activation from the previous layer, and $O_{jp}$ in the bilinear rule), making them \emph{sensitive to the precise order of computations.} This stands in contrast to the associative property found in standard arithmetic, which will cause failure for approximating LOO. 
This sensitivity to computation order hints at a deeper limitation of LRP: even in simple multiplicative settings, its attributions can depend on implementation details rather than functional behavior, which we formalize next.

\section{Why does LRP still approximate LOO poorly?}
\label{sec:theory}

While LRP's layer-local propagation rules guarantee relevance conservation, they do not guarantee consistent explanations for functionally equivalent networks. We first analyze how this axiomatic failure arises from the bilinear operations at the core of Transformer attention, formally proving that AttnLRP’s propagation rule is sensitive to the factorization of these operations. To probe the second potential source of error, we compare AttnLRP to a CP-LRP which bypasses the softmax step, allowing us to isolate softmax layer's impact from bilinear operations on LOO correlation. Taken together, these two perspectives—on bilinear factorization and softmax propagation—reveal fundamental weaknesses in current LRP variants as approximations to LOO.

\subsection{Part 1: Bilinear Propagation in LRP Violates Implementation Invariance}
\label{sec:implementation_invariance}



Implementation invariance requires that two networks computing the same function produce identical explanations \citep{Sundararajan17Axiomatic}. While methods like Integrated Gradients satisfy this axiom by design, propagation-based methods such as LRP can violate it. 
Earlier works demonstrate that LRP is not implementation invariant when applying its propagation rules to ReLU and BatchNorm layers~\citep{Sundararajan17Axiomatic,guillemot2020breaking,yeom2021pruning}. We extend this analysis to \textbf{bilinear operations}, a core component of Transformer attention, and show that the $\varepsilon$-rule for bilinear layers introduced in AttnLRP~\citep{Achtibat2024AttnLRP} are sensitive to the computational \textit{factorization} of these operations, even when the underlying function remains identical.

We demonstrate this flaw with a simple scalar example. Consider the function $f(x_1, x_2, x_3) = x_1 x_2 x_3$, which can be implemented in two computationally equivalent ways:
\begin{itemize}
\item \textbf{Model A (Left-associative):} $y = (x_1 x_2) x_3$
\item \textbf{Model B (Right-associative):} $y = x_1 (x_2 x_3)$
\end{itemize}
As illustrated in \Cref{fig:bilinear_diagram}, let $x_1 = 2, x_2 = 3, x_3 = 4$, yielding an output $y=24$. If we assign this output as the total relevance, $R_y = 24$, the standard LRP rule for multiplication (equal splitting) distributes relevance at each step.

\paragraph{Model A Derivation} The relevance is propagated backward as follows:
\begin{itemize}
\item First, the relevance $R_y = 24$ is split between $(x_1 x_2)$ and $x_3$, so each receives 12.
\item Then, the 12 assigned to $(x_1 x_2)$ is split between $x_1$ and $x_2$, so each receives 6.
\end{itemize}
This yields final relevance scores of:

$$R_{x_1} = 6,\quad R_{x_2} = 6,\quad R_{x_3} = 12$$

\paragraph{Model B Derivation} In contrast, the right-associative model's propagation is:
\begin{itemize}
\item First, the relevance $R_y = 24$ is split between $x_1$ and $(x_2 x_3)$, so each receives 12.
\item Then, the 12 assigned to $(x_2 x_3)$ is split between $x_2$ and $x_3$, so each receives 6.
\end{itemize}
This results in different final relevance scores:

$$R_{x_1} = 12,\quad R_{x_2} = 6,\quad R_{x_3} = 6$$

The relevance scores for $x_1$ and $x_3$ are swapped based on the grouping of operations, even though the function, output, and gradients are identical. In sharp contrast, the LOO scores for both implementations are identical. Setting any single input to zero makes the final output zero, causing a change of 24 from the original output ($LOO_1=24, LOO_2=24, LOO_3=24$). This is because LOO depends only on functional behavior, not on the specific parameterization.

This discrepancy not only exists in this specific example. More generally, any operation that does not satisfy the associative property can lead to implementation-dependent differences in LRP attributions. For instance, in linear attention variants that omit the softmax, the term $QK^\top V$ becomes associative and can be computed as either $(QK^\top)V$ or $Q(K^\top V)$ \citep{katharopoulos2020transformers}.
Although the softmax activation in standard attention obscures the underlying associativity, our analysis of two bilinear factorizations suggests that standard attention may also violate implementation invariance for certain mathematical operations.

\subsection{Part 2: Softmax Propagation as a Second Source of Error}
\label{sec:softmax_probe_motivation}

In addition to bilinear operations, as another key component of the attention mechanism, we further hypothesize that the propagation rules in softmax layers are also problematic. There are at least two types of error introduced by softmax propagation.

 \textbf{(1) Structural bias:} when the logits are uniform, the softmax layer outputs a nonzero, uniform attention distribution that reflects a \textit{default behavior} rather than input-dependent evidence.
LRP then spreads relevance evenly across inputs, whereas ground-truth LOO would assign near-zero relevance to each feature if all features are actually close to 0.

\textbf{(2) Linearization error:} 
As introduced in the DTD framework (\Cref{sec:loo_lrp}), 
the softmax propagation rule relies on a first-order Taylor expansion (\Cref{eq:softmax-lrp}) around the observed logits. 
When the logits are large but similar, the attention distribution appears flat even though individual inputs can strongly influence the output.
In this regime, the local Jacobian provides little discriminative signal, and the linearization fails to capture the large non-local effect of removing individual inputs---leading to systematic misallocation of relevance.

To isolate the effect of softmax from bilinear operations in LRP, we revisit \textbf{CP-LRP} as a diagnostic baseline against AttnLRP. CP-LRP treats the value readout \( O = AV \) as linear in \( V \) while holding \( A \) fixed (\Cref{sec:background}), thereby bypassing propagation through both the attention weights and, particularly, the softmax layer \( A = \softmax(Z) \). This makes CP-LRP a natural baseline for isolating the impact of softmax propagation: any gap between CP-LRP and AttnLRP on LOO reflects the added effect of the softmax rule and the bilinear split into \( A \) (\Cref{eq:attn-bilinear,eq:softmax-lrp}). 
If CP-LRP shows higher LOO agreement, it suggests that softmax propagation is a key source of attribution error.

\section{Experiments}
\label{sec:experiments}

We empirically evaluate these two error sources through three research questions:
\begin{itemize}
    \item \textbf{RQ1:} Does LRP’s $\varepsilon$-rule for bilinear layers produce different attributions for functionally equivalent factorizations (i.e., show implementation variance)?
    \item \textbf{RQ2:} If we handle the attention step like CP-LRP (send all relevance at $O$ to $V$ and skip softmax), do the attributions agree better with LOO than AttnLRP?
    \item \textbf{RQ3:} Which layers were affected the most from bypassing softmax propagation in Transformer attention as in CP-LRP?
\end{itemize}

\subsection{Experimental Setup}
\label{sec:experimental_setup}

\paragraph{Tasks and Datasets.}
We evaluate attribution methods across three complementary settings:
(1) a synthetic bilinear setting using the MNIST dataset~\citep{lecun1998mnist}, 
(2) two standard text classification benchmarks, SST~\citep{Socher2013Recursive} and IMDB~\citep{Maas2011Learning}, and 
(3) BERT-base~\citep{Devlin2019BERT} for realistic Transformer-based evaluation.  
For MNIST, we resize the images to \(14 \times 14\) and use them to probe LRP's behavior in controlled bilinear networks while retaining nontrivial real inputs.  
For SST and IMDB, we follow standard text preprocessing and fine-tune BERT-base for classification.
Implementation details are provided in \Cref{app:experiment_details}.

\paragraph{Models.}
For MNIST experiments, we design two synthetic QKV networks with linear attention without softmax layers, which differ only in how their bilinear computations are factorized (left- vs. right-associative; see \Cref{sec:implementation_invariance}). These models are lightweight but allow us to cleanly test implementation invariance on real data.  Note that real
Transformers insert a softmax layer between bilinear computations, preventing a direct associativity test. For SST and IMDB, we use a standard 12-layer BERT-base encoder. 

\paragraph{Evaluation Metrics.}
Our primary metric is the Pearson correlation ($r$) between attribution scores and ground-truth feature importance derived from \textbf{Leave-One-Out (LOO)}. As a secondary metric, we use the Area Over the Perturbation Curve (AOPC) for \textbf{Insertion and Deletion}~\citep{Petsiuk2018RISE,samek2017evaluating}, which measures how the model’s prediction changes when features are removed in order of importance.

Additionally, we compute two standard perturbation curves: 
\textbf{MoRF} (Most Relevant First) and \textbf{LeRF} (Least Relevant First)~\citep{samek2017evaluating, Petsiuk2018RISE}. 
For MoRF, features are progressively removed in decreasing order of relevance, and the model's output is recorded after each step; faithful explanations should cause rapid prediction degradation.
LeRF performs the same procedure but removes features from least to most relevant, where higher LeRF indicates more faithful explanation.
We summarize each curve using the \textit{Area Over the Perturbation Curve} (AOPC) and report LeRF, MoRF, and their difference $\Delta = \text{LeRF} - \text{MoRF}$, which reflects how well the attribution separates relevant from irrelevant features (higher is better).

\paragraph{Details for Pearson Correlation Computation and LOO.}
For each example, we compute the Pearson correlation between token- or pixel-level attribution scores and their corresponding LOO scores, then report the mean correlation across the dataset. All attribution maps are normalized to sum to one per example. To compute LOO scores, we remove each feature individually and measure the change in the model’s predicted logit. For text, this is done by masking out each token using the attention mask. For images, we zero out each pixel one at a time.

\subsection{Baselines}
\label{sec:baselines}
We evaluate \textbf{Integrated Gradients (IG)} \citep{Sundararajan17Axiomatic}, \textbf{Attention Rollout} \citep{abnar-zuidema-2020-quantifying}, \textbf{AttnLRP} \citep{Achtibat2024AttnLRP}, and \textbf{CP-LRP} \citep{Ali22cplrp}.

\subsection{Results and Analysis}

\paragraph{(RQ1) The Bilinear Rule in LRP is Not Implementation Invariant.}
To test whether LRP’s $\varepsilon$-rule for bilinear layers exhibits implementation variance, we construct two \emph{functionally equivalent} QKV linear attention networks that have two bilinear layers and differ only in the order of matrix multiplication. Each consists of three learned projections ($W_Q$, $W_K$, $W_V$), followed by two matrix multiplications and a linear output layer. Neither includes softmax, so the bilinear operation is associative. The only difference is whether the product is evaluated right-associatively $(QK^\top)V$ or left-associatively $Q(K^\top V)$ as we proposed at the end of \cref{sec:implementation_invariance}. 
Both networks are trained on $14\times14$ MNIST (~85\% accuracy) with shared weights, ensuring identical functions but different computational graphs.

We compute attributions using LOO, IG, and AttnLRP, and measure Pearson correlations between left and right models and with LOO (\Cref{tab:bilinear_implinv_results}). LOO is perfectly invariant ($r=1.0$), and IG also yields perfect left–right agreement, consistent with its axiomatic invariance. AttnLRP, however, is implementation dependent ($r=0.79$). Compared to LOO, IG shows a negative but nonzero correlation ($r=-0.37$), whereas AttnLRP exhibits near-zero correlation ($r\approx 0.07$ left; $r\approx -0.07$ right), indicating both implementation dependence and poor alignment. 
These results show that implementation invariance is necessary but not sufficient: IG passes the invariance test but does not reliably align with LOO---sometimes exhibiting negative or later shown near-zero correlations---while AttnLRP fails on both.
Additional analysis on formulation of IG is in \Cref{app:ig_comparison}.

\paragraph{(RQ2) Bypassing Softmax Improves Alignment with LOO.}
To isolate the effect of softmax propagation in AttnLRP, we compare its performance with CP-LRP. As shown in \Cref{tab:baselines_sst_imdb}, CP-LRP substantially improves correlation with LOO on SST over AttnLRP (e.g., \( r = 0.52 \) on SST versus \( r = 0.22 \) for AttnLRP). Improvements on IMDB are smaller, potentially because its longer sentences lead to more correlated features. These findings support the claim that the softmax propagation rule is a major source of attribution error.



\begin{table}[t]
    \centering
    
    \caption{\textbf{LRP fails implementation invariance and alignment with LOO in bilinear layers.} We compare feature attributions between left- and right-associative QKV bilinear networks that compute the same function. LOO and IG exhibit perfect implementation invariance (L vs R = 1), whereas AttnLRP does not. Correlations with LOO are computed separately for the left and right models (identical for LOO by definition), showing that IG has negative but nontrivial correlation whereas AttnLRP is near zero, indicating that invariance alone is not sufficient for faithfulness.}
    \label{tab:bilinear_implinv_results}
    \small
    \begin{tabular}{lccc}
    \toprule
    \textbf{Explainer} & \textbf{L vs R} & \textbf{L vs LOO} & \textbf{R vs LOO} \\
    \midrule
    LOO      & 1.0000 & 1.0000 & 1.0000 \\
    IG  & 1.0000 & $-0.3707$ & $-0.3707$ \\
    AttnLRP  & \underline{0.7865} & \underline{$0.0730$}  & \underline{$-0.0698$} \\
    \bottomrule
    \end{tabular}
\end{table}

\begin{table}[t]
\centering
\caption{Baselines on attribution metrics (SST top, IMDB bottom). 
CP\textnormal{-}LRP shows substantially higher agreement with LOO than AttnLRP, 
highlighting the softmax propagation rule as a source of error.
Metrics: LOO~$r$, LeRF, MoRF, and $\Delta=\text{LeRF}-\text{MoRF}$.
First place is \textbf{bolded} and second place is \textit{italicized} (LOO’s correlation with itself excluded).}
\label{tab:baselines_sst_imdb}
\small
\setlength\tabcolsep{4pt}
\begin{tabular}{@{}lcccc@{}}
\toprule
Method & LOO~$r$ $\uparrow$ & LeRF $\uparrow$ & MoRF $\downarrow$ & $\Delta \uparrow$ \\
\midrule
\textbf{SST} \\
LOO & 1.00 & \textbf{1.11} & 0.33 & 0.78 \\
IG & -0.05 & 0.48 & 0.36 & 0.12 \\
AttentionRollout & 0.08 & 0.72 & 0.38 & 0.35 \\
\addlinespace[2pt]
\multicolumn{5}{@{}l}{\textsc{LRP variants}}\\[-0.6ex] 
\cmidrule(lr){1-5}
AttnLRP & \textit{0.22} & 0.88 & \textbf{0.05} & \textbf{0.83} \\
CP\textnormal{-}LRP & \textbf{0.52} & \textit{1.00} & \textit{0.22} & \textit{0.79} \\
\midrule
\textbf{IMDB} \\
LOO & 1.00 & 2.20 & -0.94 & 3.14 \\
IG & 0.00 & 1.98 & 2.70 & -0.72 \\
AttentionRollout & 0.04 & 2.85 & 1.38 & 1.47 \\
\addlinespace[2pt]
\multicolumn{5}{@{}l}{\textsc{LRP variants}}\\[-0.6ex] 
\cmidrule(lr){1-5}
AttnLRP & \textit{0.25} & \textbf{3.72} & \textit{-2.64} & \textit{6.36} \\
CP\textnormal{-}LRP & \textbf{0.26} & \textbf{3.73} & \textbf{-3.02} & \textbf{6.75} \\
\bottomrule
\end{tabular}
\end{table}

\begin{figure*}[t]
  \centering
  \begin{minipage}[t]{0.49\textwidth}
    \centering
    \includegraphics[width=\linewidth]{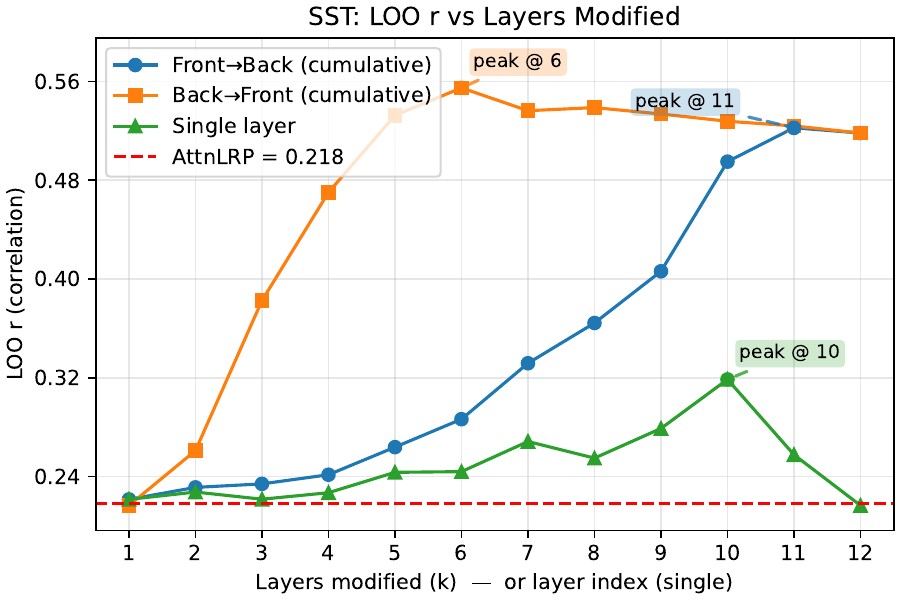}
    \subcaption{SST LOO $r$}\label{fig:a}
  \end{minipage}\hfill
  \begin{minipage}[t]{0.49\textwidth}
    \centering
    \includegraphics[width=\linewidth]{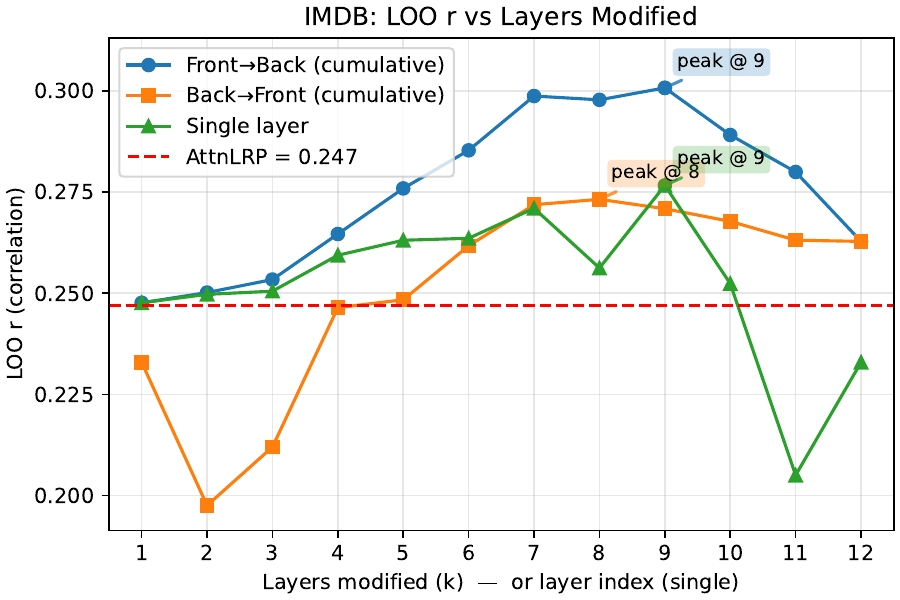}
    \subcaption{IMDB LOO $r$}\label{fig:b}
  \end{minipage}

  \vspace{1.4em}

  \begin{minipage}[t]{0.49\textwidth}
    \centering
    \includegraphics[width=\linewidth]{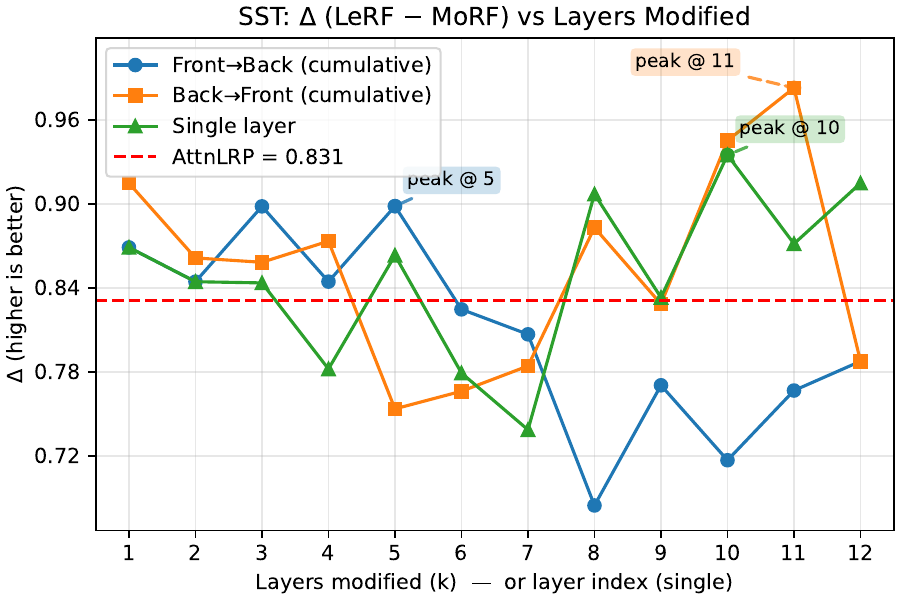}
    \subcaption{SST $\Delta=\text{LeRF}-\text{MoRF}$}\label{fig:c}
  \end{minipage}\hfill
  \begin{minipage}[t]{0.49\textwidth}
    \centering
    \includegraphics[width=\linewidth]{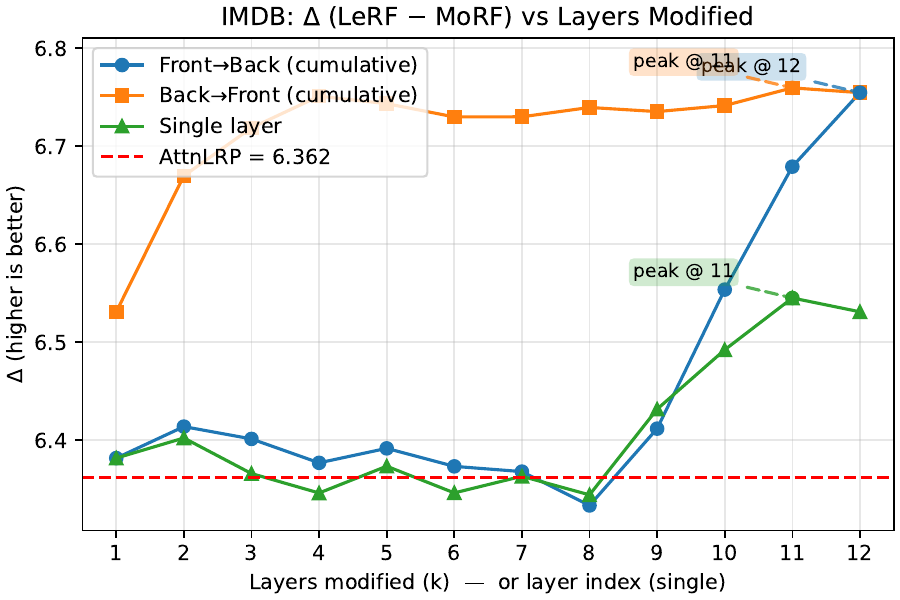}
    \subcaption{IMDB $\Delta=\text{LeRF}-\text{MoRF}$}\label{fig:d}
  \end{minipage}

  \vspace{0.8em}
\caption{
\textbf{Layerwise impact of bypassing softmax with CP-LRP on attribution faithfulness.}
We compare Pearson correlation with LOO ($r$, higher is better) and the AOPC difference ($\Delta=\text{LeRF}-\text{MoRF}$, higher is better) on SST and IMDB. 
We ablate the softmax propagation rule by removing it from: the first $k$ layers (Front-to-Back), the last $k$ layers (Back-to-Front), or only a single layer. 
\textbf{The dashed red line indicates the performance of the standard AttnLRP baseline; points above this line show an improvement in faithfulness.} 
The largest gains appear in several middle and later layers, consistent across both metrics.
For complete numerical results, see Table~\ref{tab:layer_softmax_full_sst_imdb} in the appendix.}
  \label{fig:layer_softmax_comparison}
\end{figure*}

\paragraph{(RQ3) Identifying the most impacted layers.}
We localize where softmax propagation contributes most by applying the CP-LRP in three ways: (1) a single layer at a time, (2) cumulatively from the first $k$ layers (front-to-back), and (3) cumulatively from the last $k$ layers (back-to-front). 
\Cref{fig:layer_softmax_comparison} shows that correlation with LOO improves most when modifying \textbf{middle and later layers}. 
Note that applying the identity at \emph{all} layers coincides with CP-LRP’s behavior at the attention step.


On SST, for example, the ``Single layer'' ablation peaks around layers 6 and 10, while the ``Front-to-Back'' curve shows a steep rise starting from layer 6. These results indicate that the softmax rule introduces the most attribution error in the model’s middle-to-late layers, where feature interactions are more complex.


\paragraph{Summary of Findings.}
Our experiments yield three takeaways.
(1) In a controlled bilinear setting without softmax, AttnLRP’s $\varepsilon$-rule for bilinear layers violates implementation invariance, assigning different attributions to functionally identical factorizations.
(2) The softmax step is a major source of attribution error in Transformers: CP-LRP achieves higher agreement with LOO where on SST, $r$ rises from 0.22 (AttnLRP) to 0.52 (CP-LRP).
(3) Layer-wise ablations show that gains concentrate in middle-to-late layers, indicating where softmax propagation harms faithfulness most.

\section{Related Work}
\label{sec:related}


\paragraph{Approximating Feature Importance.}
Much research has sought efficient approximations for the computationally prohibitive Leave-One-Out (LOO) method. 
Early heuristics such as Attention Rollout \citep{abnar-zuidema-2020-quantifying} offered intuitive ways to trace information flow through attention layers, but attention weights themselves have been challenged as faithful explanations \citep{jain-wallace-2019-attention}, with follow-up work clarifying conditions under which they can be informative \citep{wiegreffe-pinter-2019-attention}. 
More principled approaches based on relevance conservation, most notably Layer-Wise Relevance Propagation (LRP) \citep{Bach2015LRP}, provide efficient single-pass alternatives, spurring adaptations for Transformers \citep{Ali22cplrp} and extensions to bilinear settings and attention mechanisms, including BiLRP for dot-product similarity models \citep{eberle2020building} and AttnLRP for Transformers \citep{Achtibat2024AttnLRP}. 
These methods focus on deriving propagation rules and have shown promise as efficient substitutes for LOO, but their axiomatic properties remain underexplored—especially in modern architectures with complex bilinear interactions.

\paragraph{Implementation Invariance and Canonization.}
The reliability of attribution methods is often judged by formal axioms \citep{Sundararajan17Axiomatic}, among which \textit{implementation invariance} plays a central role: explanations should depend only on a model’s function, not its specific parameterization~\citep{kindermans2019reliability}. 
While methods like Integrated Gradients (IG) satisfy this axiom by design, propagation-based methods such as LRP \citep{Montavon2018Methods,shrikumar2017learning} do not. 
\citet{Sundararajan17Axiomatic} first illustrated this flaw by rearranging ReLU nonlinearities in functionally equivalent networks, leading to different LRP attributions. 
Subsequent works find similar violations in BatchNorm layers and solve it with \emph{model canonization}: merging BatchNorm with preceding convolutions to stabilize explanations and reduce implementation-dependent artifacts~\citep{guillemot2020breaking,yeom2021pruning}.
While effective for CNNs, these techniques do not address the core propagation rules for bilinear and softmax layers in Transformers, which rely on LayerNorm instead of BatchNorm.


\paragraph{Perspectives on Bilinear Layers.}
Bilinear layers, a core component of attention, have been studied from both axiomatic and mechanistic perspectives. 
From an axiomatic standpoint, recent theoretical results show that no attribution method assigning relevance to individual features can faithfully explain polynomial functions with correlated inputs, motivating group-based attributions in bilinear settings \citep{you2025sumofparts}. 
Mechanistic interpretability work treats bilinear operations as the structural backbone of ``attention circuits,'' enabling reverse-engineering of model computations \citep{elhage2021mathematical,Nanda2023Progress}. 
Bilinear layers have also been shown to admit linear tensor decompositions that expose pairwise interactions, making them mathematically tractable for analysis \citep{sharkey2023technicalnotebilinearlayers}.
By contrast, our work focuses on \emph{implementation invariance}, formally proving that LRP’s propagation rules for bilinear layers violate this axiom even in simplified settings.

\paragraph{Evaluating Explanations.}
Evaluating explanations involves multiple desiderata that capture different aspects of explanatory quality, since no single metric suffices \citep{Jacovi2021Formalizing,Atanasova2023Faithfulness}. 
Key desiderata include \emph{faithfulness}, \emph{stability and consistency}, \emph{structural properties}, and \emph{expert alignment}.
\textit{Faithfulness} assesses how well explanations reflect the model’s behavior. Standard approaches include sanity checks \citep{Adebayo2018Sanity}, retraining-based benchmarks \citep{Hooker2019A}, and post-hoc metrics such as Leave-One-Out (LOO) and insertion/deletion curves \citep{Lundberg2017A,Petsiuk2018RISE,samek2017evaluating,Atanasova2023Faithfulness,feng-etal-2018-pathologies}. 
\textit{Stability} assesses the robustness of explanations to perturbations in explanations or inputs \citep{slack2021counterfactual,xue2023stability,kim2024evaluating,jin2025probabilistic,you2025probabilistic}.
\textit{Structural properties}, such as contiguity or sparsity, evaluate whether explanations form coherent and interpretable patterns rather than fragmented noise \citep{kim2024evaluating,you2025sumofparts}. 
Beyond these, other work has proposed causal or environment-invariant criteria for explanations, aiming to identify rationales that remain predictive across different environments \citep{Chang2019Invariant}.
\textit{Expert alignment} measures agreement with human or domain-expert expectations \citep{doshi2017towards,jin2025fix,Havaldar2025TFIX,Lage2019HumanEvaluation,nguyen-2018-comparing}.
In this work, we focus on efficiently approximating LOO and complement it with standard perturbation-based metrics.

\section{Conclusion}
We study when LRP-style attributions align with Leave-One-Out (LOO) in Transformers and identify two key mismatches: 
(1) bilinear rules in AttnLRP violate implementation invariance, and 
(2) softmax propagation introduces linearization errors. 
Empirically, bypassing softmax and propagating only through values, as in CP-LRP, yields better LOO alignment. 
A promising fix is to canonize larger attention blocks during relevance propagation, 
reducing both bilinear implementation variance and softmax linearization errors—analogous to merging BatchNorm with preceding layers in CNNs. 
Such block-wise propagation may offer a more faithful approximation to LOO and guide future work on efficient, theoretically grounded attribution methods.

\section*{Limitations}
This study is limited to a toy analytic example, BERT, and a simple linear attention network; generalization to other attention-based architectures remains to be explored. Our analysis focuses on attention layers under a specific set of LRP design choices, and does not exhaustively compare alternative propagation rules or gradient-based attribution methods. Our LOO reference is based on masking-based removal, which may induce distribution shift. 
Evaluation primarily relies on Pearson correlation with LOO and perturbation-based metrics, though other evaluation metrics exist.
Finally, while we identify several failure modes of current LRP variants, we do not propose a remedy; developing axiomatically grounded and efficient alternatives is left for future work.
\section*{Acknowledgments}
Part of this work was done when Weiqiu You and Siqi Zeng were visiting research students at Okinawa Institute of Science and Technology.
WY was supported by a gift from AWS AI to Penn Engineering's ASSET Center for Trustworthy AI. SZ and HZ were partly supported by an NSF IIS grant No.\ 2416897 and an NSF CAREER Award No.\ 2442290. HZ would like to thank the support of a Google Research Scholar Award and Nvidia Academic Grant Award. MY was partly supported by JSPS KAKENHI Grant Number 24K03004 and by JST ASPIRE JPMJAP2302. The views and conclusions expressed in this paper are solely those of the authors and do not necessarily reflect the official policies or positions of the supporting companies and government agencies.

\bibliography{main}

\appendix

\renewcommand{\thetable}{A\arabic{table}}
\renewcommand{\thefigure}{A\arabic{figure}}

\begin{table*}[t]
\centering
\caption{Layer‐wise effects for bypassing softmax in backpropagation  on attribution metrics (SST top, IMDB bottom). Metrics: LOO~$r$, LeRF, MoRF, and $\Delta=\text{LeRF}-\text{MoRF}$. $L_{:3}$ means removing layers 1-3. $L_{7:}$ means removing layers 7-12.}
\label{tab:layer_softmax_full_sst_imdb}
\small
\setlength\tabcolsep{3pt}
\begin{tabular}{@{}lcccc | lcccc | lcccc@{}}
\toprule
\multicolumn{5}{c|}{\textbf{Front $\rightarrow$ Back}} & \multicolumn{5}{c|}{\textbf{Back $\rightarrow$ Front}} & \multicolumn{5}{c}{\textbf{Single Layer}} \\[0.25ex]
\cmidrule(lr){1-5} \cmidrule(lr){6-10} \cmidrule(lr){11-15}
Rm Layers & LOO~$r$ & LeRF & MoRF & $\Delta$ & Rm Layers & LOO~$r$ & LeRF & MoRF & $\Delta$ & Rm Layer & LOO~$r$ & LeRF & MoRF & $\Delta$ \\
\midrule
\textbf{SST} \\
$L_{:1}$ & 0.22 & 0.92 & 0.05 & 0.87 & $L_{12:}$ & 0.22 & 0.89 & -0.02 & 0.91 & $L_{1}$ & 0.22 & 0.92 & 0.05 & 0.87 \\
$L_{:2}$ & 0.23 & 0.89 & 0.05 & 0.84 & $L_{11:}$ & 0.26 & 0.91 & 0.05 & 0.86 & $L_{2}$ & 0.23 & 0.89 & 0.05 & 0.84 \\
$L_{:3}$ & 0.23 & 0.89 & -0.01 & 0.90 & $L_{10:}$ & 0.38 & 0.90 & 0.04 & 0.86 & $L_{3}$ & 0.22 & 0.88 & 0.04 & 0.84 \\
$L_{:4}$ & 0.24 & 0.90 & 0.05 & 0.84 & $L_{9:}$ & 0.47 & 0.97 & 0.10 & 0.87 & $L_{4}$ & 0.23 & 0.84 & 0.06 & 0.78 \\
$L_{:5}$ & 0.26 & 0.88 & -0.02 & 0.90 & $L_{8:}$ & 0.53 & 0.93 & 0.18 & 0.75 & $L_{5}$ & 0.24 & 0.91 & 0.05 & 0.86 \\
$L_{:6}$ & 0.29 & 0.88 & 0.06 & 0.82 & $L_{7:}$ & 0.55 & 0.97 & 0.20 & 0.77 & $L_{6}$ & 0.24 & 0.90 & 0.12 & 0.78 \\
$L_{:7}$ & 0.33 & 0.82 & 0.02 & 0.81 & $L_{6:}$ & 0.54 & 0.96 & 0.18 & 0.78 & $L_{7}$ & 0.27 & 0.89 & 0.15 & 0.74 \\
$L_{:8}$ & 0.36 & 0.82 & 0.14 & 0.68 & $L_{5:}$ & 0.54 & 1.06 & 0.17 & 0.88 & $L_{8}$ & 0.25 & 0.96 & 0.05 & 0.91 \\
$L_{:9}$ & 0.41 & 0.91 & 0.14 & 0.77 & $L_{4:}$ & 0.53 & 1.01 & 0.18 & 0.83 & $L_{9}$ & 0.28 & 0.93 & 0.09 & 0.83 \\
$L_{:10}$ & 0.50 & 0.85 & 0.14 & 0.72 & $L_{3:}$ & 0.53 & 1.11 & 0.16 & 0.95 & $L_{10}$ & 0.32 & 0.93 & -0.01 & 0.93 \\
$L_{:11}$ & 0.52 & 0.94 & 0.18 & 0.77 & $L_{2:}$ & 0.52 & 1.12 & 0.13 & 0.98 & $L_{11}$ & 0.26 & 0.91 & 0.04 & 0.87 \\
$L_{:12}$ & 0.52 & 1.00 & 0.22 & 0.79 & $L_{1:}$ & 0.52 & 1.00 & 0.22 & 0.79 & $L_{12}$ & 0.22 & 0.89 & -0.02 & 0.91 \\
\addlinespace
\midrule
\textbf{IMDB} \\
$L_{:1}$ & 0.25 & 3.72 & -2.66 & 6.38 & $L_{12:}$ & 0.23 & 3.72 & -2.81 & 6.53 & $L_{1}$ & 0.25 & 3.72 & -2.66 & 6.38 \\
$L_{:2}$ & 0.25 & 3.72 & -2.69 & 6.41 & $L_{11:}$ & 0.20 & 3.72 & -2.95 & 6.67 & $L_{2}$ & 0.25 & 3.72 & -2.68 & 6.40 \\
$L_{:3}$ & 0.25 & 3.73 & -2.68 & 6.40 & $L_{10:}$ & 0.21 & 3.72 & -3.00 & 6.72 & $L_{3}$ & 0.25 & 3.72 & -2.64 & 6.37 \\
$L_{:4}$ & 0.26 & 3.73 & -2.65 & 6.38 & $L_{9:}$ & 0.25 & 3.73 & -3.02 & 6.75 & $L_{4}$ & 0.26 & 3.72 & -2.63 & 6.35 \\
$L_{:5}$ & 0.28 & 3.73 & -2.67 & 6.39 & $L_{8:}$ & 0.25 & 3.72 & -3.02 & 6.74 & $L_{5}$ & 0.26 & 3.72 & -2.65 & 6.37 \\
$L_{:6}$ & 0.29 & 3.72 & -2.65 & 6.37 & $L_{7:}$ & 0.26 & 3.72 & -3.01 & 6.73 & $L_{6}$ & 0.26 & 3.72 & -2.63 & 6.35 \\
$L_{:7}$ & 0.30 & 3.73 & -2.64 & 6.37 & $L_{6:}$ & 0.27 & 3.73 & -3.00 & 6.73 & $L_{7}$ & 0.27 & 3.72 & -2.64 & 6.36 \\
$L_{:8}$ & 0.30 & 3.72 & -2.62 & 6.33 & $L_{5:}$ & 0.27 & 3.73 & -3.01 & 6.74 & $L_{8}$ & 0.26 & 3.72 & -2.63 & 6.34 \\
$L_{:9}$ & 0.30 & 3.73 & -2.69 & 6.41 & $L_{4:}$ & 0.27 & 3.73 & -3.00 & 6.74 & $L_{9}$ & 0.28 & 3.72 & -2.71 & 6.43 \\
$L_{:10}$ & 0.29 & 3.74 & -2.81 & 6.55 & $L_{3:}$ & 0.27 & 3.73 & -3.01 & 6.74 & $L_{10}$ & 0.25 & 3.73 & -2.77 & 6.49 \\
$L_{:11}$ & 0.28 & 3.74 & -2.94 & 6.68 & $L_{2:}$ & 0.26 & 3.74 & -3.02 & 6.76 & $L_{11}$ & 0.21 & 3.72 & -2.82 & 6.55 \\
$L_{:12}$ & 0.26 & 3.73 & -3.02 & 6.75 & $L_{1:}$ & 0.26 & 3.73 & -3.02 & 6.75 & $L_{12}$ & 0.23 & 3.72 & -2.81 & 6.53 \\
\bottomrule
\end{tabular}
\end{table*}

\section{Comparison with Integrated Gradients}
\label{app:ig_comparison}

Integrated Gradients (IG) is an axiomatically-sound method that satisfies implementation invariance by design \citep{Sundararajan17Axiomatic}. It computes feature attributions by integrating gradients along a path from a baseline input \(x'\) to the actual input \(x\):
\[
\mathrm{IG}_i(x) = (x_i - x'_i) \int_0^1 \frac{\partial f(x' + \alpha(x - x'))}{\partial x_i} \, d\alpha
\]
Because IG's attribution depends only on the function \(f\) and not the network's specific architecture, it would produce identical explanations for both the scalar and attention counterexamples discussed in Section~\ref{sec:theory}, regardless of how the operations are grouped. While IG is computationally more intensive, this contrast highlights a fundamental trade-off: propagation-based methods like LRP offer efficiency but can fail to preserve fundamental axioms.

\section{Experiment Details}
\label{app:experiment_details}



\subsection{Hyperparameters}
\subsubsection{MNIST Linear Attention Experiments}
For MNIST, we train simple models \texttt{QKVLeftAssoc\_Seq\_NoSoftmax} and \texttt{QKVRightAssoc\_Seq\_NoSoftmax}, which are linear attention networks without softmax normalization. Input images are resized from $28 \times 28$ to $14 \times 14$ and flattened into sequences of length $196$. Both models use an embedding dimension of $32$ and predict over $10$ classes. We train each model for 5 epochs using Adam (learning rate $0.001$, $\beta_1=0.9$, $\beta_2=0.999$, $\epsilon=10^{-8}$) with cross-entropy loss. 

For LOO, we remove \emph{single pixels} on the $14 \times 14$ resized images by setting them to zero and measuring the change in the predicted class logit (batch size 8 for pixel sweeps). Attribution is computed with LRP-$\varepsilon$ ($\varepsilon=10^{-6}$), and attributions are normalized to sum to 1 before computing Pearson correlation with LOO, complemented by MoRF/LeRF perturbation curves.

\subsubsection{Text Experiments}
For text experiments, we use pretrained BERT-base models from Hugging Face. For SST, we use \texttt{textattack/bert-base-uncased-SST-2}, and \texttt{fabriceyhc/bert-base-uncased-imdb} for IMDB. Both models are standard BERT-base architectures fine-tuned for sentiment classification.

\paragraph{LOO granularity.} For both SST and IMDB, LOO is computed at the \textbf{single-token} level by masking one token at a time and measuring the change in the predicted class logit.

\paragraph{MoRF/LeRF granularity.} For IMDB only, MoRF and LeRF perturbation curves are computed over \textbf{non-overlapping contiguous 16-token chunks} to make perturbation more computationally tractable on long sequences. For SST, MoRF and LeRF use \textbf{single-token} removals. This grouping affects only the perturbation curves; all LOO–based correlations use single-token LOO.

\subsection{Full Results}
Table~\ref{tab:layer_softmax_full_sst_imdb} provides the full numerical results for our layer-wise softmax ablation study, corresponding to the data visualized in Figure~\ref{fig:layer_softmax_comparison} in the main text.

\end{document}